\documentclass[letterpaper, 10 pt, conference]{packages/ieeeconf} 
\IEEEoverridecommandlockouts
          \usepackage[compress]{cite}
\makeatletter
\let\NAT@parse\undefined
\makeatother
\usepackage[colorlinks=true]{hyperref}
\usepackage{arydshln} \usepackage[skip=2pt,font=footnotesize]{caption} \usepackage{upgreek,bm}
\usepackage{amsmath, amssymb, amsfonts}
\DeclareMathSymbol{\shortminus}{\mathbin}{AMSa}{"39}
\usepackage{graphicx}
\usepackage{adjustbox}
\usepackage{subcaption}
\let\labelindent\relax
\usepackage{enumitem} \usepackage{booktabs} \usepackage{colortbl}
\usepackage{float} \usepackage{color} \usepackage[super]{nth}
\usepackage{stmaryrd}  \allowdisplaybreaks
\newcommand\scalemath[2]{\scalebox{#1}{\mbox{\ensuremath{\displaystyle #2}}}}

\usepackage{xcolor}
\usepackage{multirow}
\usepackage{xfrac}
\usepackage{todonotes}

\newcommand\gb[1]{\textbf{\color{green!80!black}{#1}}} \newcommand\gs[1]{\textbf{\textcolor{blue}{#1}}} 

\title{\LARGE \bf
Distributed Visual-Inertial Cooperative Localization
}
\author{
Pengxiang Zhu*,
Patrick Geneva*, Wei Ren, and Guoquan Huang
\thanks{*These authors contributed equally to this work.}
\thanks{
Zhu and Ren are with the Department of Electrical and Computer Engineering, University of California, Riverside, CA 92521, USA.
Email: pzhu008@ucr.edu, ren@ee.ucr.edu.
Geneva and Huang are with the Robot Perception and Navigation Group (RPNG), University of Delaware, Newark, DE 19716, USA. 
Email: \{pgeneva, ghuang\}@udel.edu}
\thanks{
This work was partially supported by 
the University of Delaware (UD) College of Engineering, 
the NSF (IIS-1924897, CMMI-2027139), and the ARL (W911NF-19-2-0226). }
}

\allowdisplaybreaks \renewcommand{\baselinestretch}{0.97}
\begin{document}
\maketitle
\thispagestyle{empty}
\pagestyle{empty}

\begin{abstract}

In this paper we present a consistent and distributed state estimator for multi-robot cooperative localization (CL)
which efficiently fuses environmental features and loop-closure constraints across time {\em and} robots.
In particular, 
we leverage covariance intersection (CI) to allow each robot to only estimate its own state and autocovariance and compensate for the unknown correlations between robots.
{Two novel multi-robot methods for utilizing common environmental SLAM features are introduced and evaluated in terms of accuracy and efficiency.}
Moreover, we adapt CI to enable drift-free estimation through the use of loop-closure measurement constraints to other robots' historical poses without a significant increase in computational cost.
The proposed distributed CL estimator is validated against its non-realtime centralized counterpart extensively in both simulations and real-world experiments.

\end{abstract}

 \section{Introduction}

Camera and inertial measurement unit (IMU) pairs have been at the forefront of multi-robot (or mobile device) applications due to their complementary nature, low cost and small size. 
Accurate and efficient cooperative localization (CL) that enables  multi-user augmented reality (AR) experiences, multi-device cooperative mapping, and multi-vehicle formation control, 
is a key barrier to overcome
due to challenges of communication, distributed computation, and complexity of multi-robot asynchronous measurement constraints.

In this paper, building upon our recent work~\cite{Zhu2021ICRA},
we propose a fully distributed multi-robot visual-inertial CL estimator
by delicately exploiting information contained in both {environmental SLAM landmarks} and loop-closures across robots and time.
Specifically, we extend our prior CI-based cooperative visual-inertial odometry (VIO) system~\cite{Zhu2021ICRA} 
to include both {SLAM features} and incorporate loop-closure constraints to historical states of other robots,
thus limiting the current robot's localization drift essentially using {\em asynchronous} common views seen from other robots' historical poses.
As a result, the proposed distributed CL estimator does {\em not} require {\em simultaneous} viewing of the same location 
due to leveraging of historical common features (e.g., a robot can gain information if another robot had previously explored the same location),
while significantly improving the localization performance thanks to such common multi-robot measurement information.
In summary, the main contributions of this work are the following:
\begin{itemize}
\item We develop a fully distributed CI-based visual-inertial CL estimation algorithm,
which allows for accurate, efficient and consistent estimation of all robot states.
\item We propose two different {SLAM feature} measurement models that allow for cooperative estimation of common long-lived environmental features, 
and validate their relative accuracy and computational complexity through a series of simulations.
\item We introduce a computationally efficient method for long-term loop-closure to reduce localization drift, which enables multi-robot constraints between historical poses and features,
allowing for robots to gain additional constraints even in the case when other robots are not actively in the same location. 
\item We thoroughly validate the proposed approach in Monte Carlo simulations and real world experiments by comparing to centralized CL algorithms.
\end{itemize}

\begin{figure}[t]
\centering
\includegraphics[trim=0 0 0 0,clip,width=0.80\columnwidth]{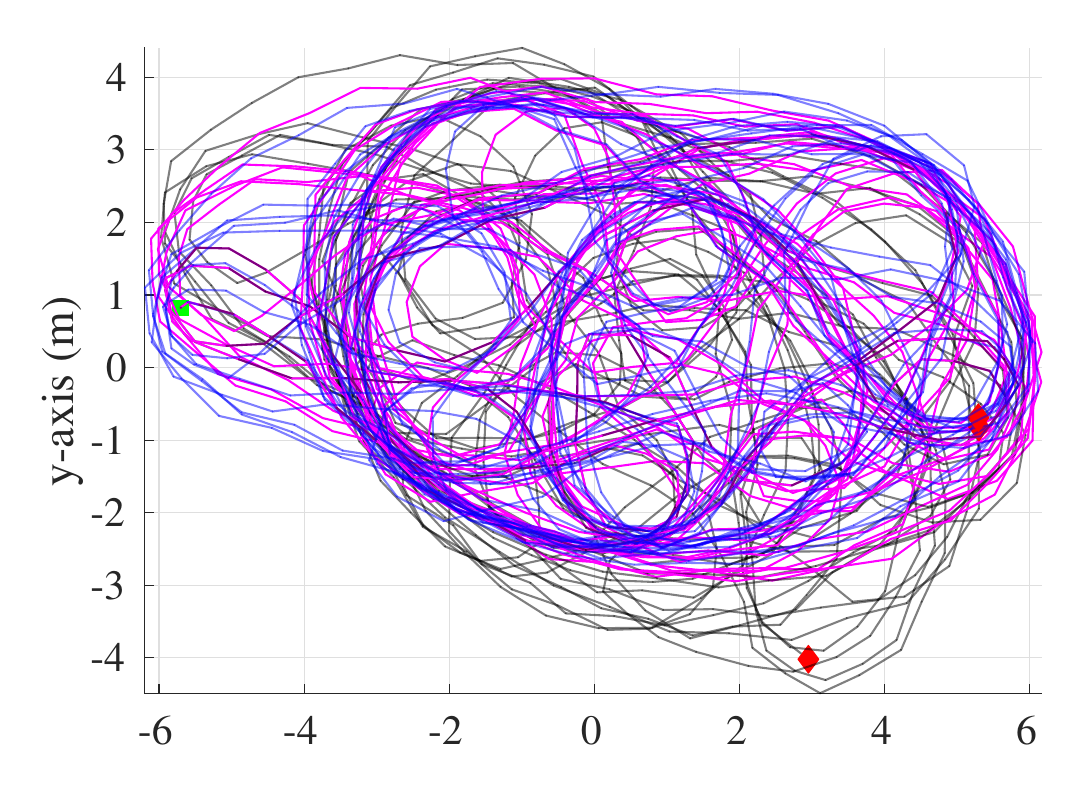}
\includegraphics[trim=0 0 0 0.5cm,clip,width=0.80\columnwidth]{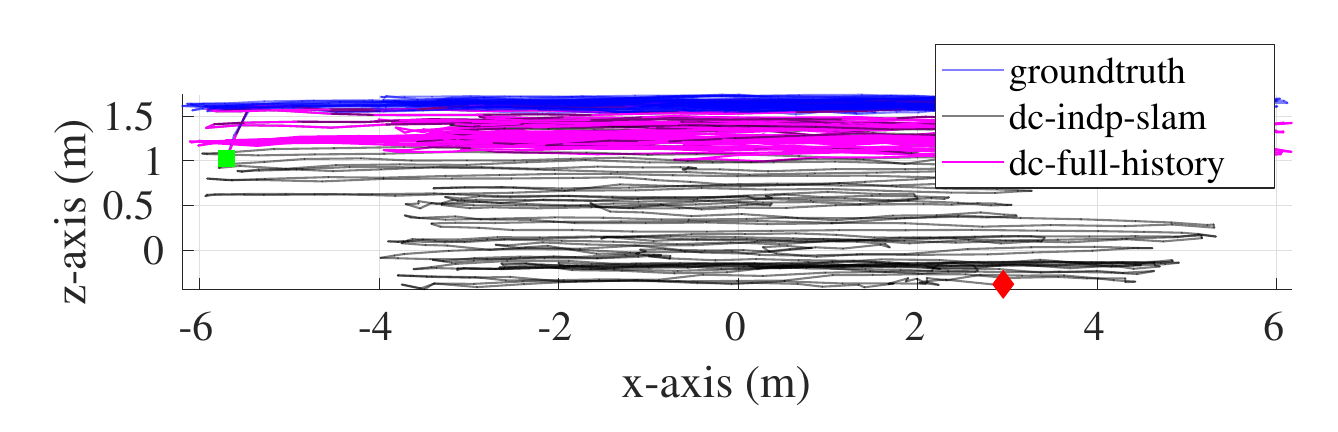}
\caption{
Trajectory of groundtruth, independent, and distributed historical trajectory for Robot 0 in the Vicon room dataset.
It can be seen that the use of common historical features limit drift in the z-axis along with improvements in x-y accuracy.
Please refer to the color figure.
}
\label{fig:loop_path}
\vspace*{-1.5em}
\end{figure}

\section{Related Work}

Significant research efforts have recently been devoted to visual-inertial navigation system (VINS) \cite{Huang2019ICRA}, 
while primarily focusing on improving {\em single-robot} VINS accuracy, efficiency, and robustness \cite{Mourikis2007ICRA,Qin2018TRO,Geneva2019CVPR}.
The extension to the {\em multi-robot} case is not sufficiently explored as a naive approach would be prohibitively costly and non-realtime.
For example, one could communicate all measurements generated from itself to each other (or fusion center), where all measurements could be optimally fused and all states can be refined jointly.
While this does allow for accurate estimation, both the requirement for constant communication and the joint estimation of robot states requires cubic computational complexity in terms of the number of robots.
As such, a multi-robot \textit{distributed} estimator is needed to address these shortcomings by relaxing communication requirements and distributing the computation cost across all robots.

Efficient 2D CL has focused on the fusion of relative measurements between robots  (e.g., relative robot-to-robot bearing or distance range measurements).
Roumeliotis et al. \cite{Roumeliotis2002TRA} proposed a decentralized algorithm 
that achieves performance equivalent to the centralized formulation, but required communication between all robots and increases in computational cost due to its centralized nature as the number of robots grow.
Other works such as~\cite{Luft2018IJRR} have investigated the approximation of the robot-to-robot cross-covariances that are not involved in a relative measurement update to reduce the computational cost, and while it
performs close to its centralized, it is unable to guarantee consistency and thus can easily diverge.
More recently, Jung et al. \cite{Jung2020ICRA} extended this work to the 3D case,
but inherits the same underlying issues and requires maintaining of the approximated robot-to-robot cross-covariances.
There exist other works aiming at estimating the relative poses between robots using relative measurements \cite{Martinelli2019RAL,Xu2020ICRA}. 
Alternative approaches have leveraged CI~\cite{Carrillo2013IROS, Zhu2020TCST} to guarantee  consistency and only requires that each robot maintains its own state and auto-covariance (the correlations between robots are ignored).
By contrast, in this work we specifically take advantage of the CI formulation for 3D multi-robot state estimation, enabling a consistent distributed algorithm which fuses inertial and visual sparse environmental feature information.

As compared to CL with relative distance, bearing, or poses between robots \cite{Roumeliotis2002TRA, Luft2018IJRR, Leung2011IJRR, Carrillo2013IROS, Zhu2020TCST, Jung2020ICRA,Martinelli2019RAL, Martinelli2020RAL, Lajoie2020RAL, Xu2020ICRA}, 
common sparse environmental features are used in~\cite{Paull2015ICRA,Karrer2018RAL,Melnyk2012ICRA,Sartipi2019IROS},
which is appealing  as getting relative robot information can be difficult with visual-inertial sensors in practice and requires both the detection and tracking of other robots.
For example, Melnyk et al. \cite{Melnyk2012ICRA} introduced CL-MSCKF using common environmental feature constraints within a centralized formulation that jointly estimated all robot states.
They required that robots communicate all sensor data to a common fusion center and demonstrated its use for the two robot case in simulation.
Karrer et al. \cite{Karrer2018RAL}  developed a graph-based centralized server which handled non-realtime computationally expensive loop closure detection and optimization of all robot maps to find the joint global optimal.
In this paper, we instead focus on the computationally efficient \textit{distributed} localization problem where each robot only estimates its \textit{own} state and tries to leverage information from other robots without a centralized server or joint optimization.

As closest to our work, 
Sartipi et al. \cite{Sartipi2019IROS} introduced a distributed method for multi-user AR experiences through the use of multi-map feature constraints.
Common features were detected in environmental maps received from other users and the transmitted feature position estimates were used to constrain the user's state directly.
Instead of inflating measurement noise to compensate for the unknown correlations between the current user and the other user's map, we leverage CI {that theoretically guarantee consistency} to handle the unknown correlations.
Also, instead of requiring that all common features must match to sparse features in the other user's map, 
we leverage the other user's common feature measurements directly allowing for update with additional measurements.

 \section{Cooperative Visual-Inertial System}

In this section, we briefly describe the cooperative visual-inertial system that serves the basis for the proposed distributed CI-based estimator.
The state vector for the  $i$'th robot contains its current IMU navigation state $\mathbf{x}_{I_i}$, sliding window of cloned IMU poses $\mathbf{x}_{C_i}$, spatial-temporal calibration parameters $\mathbf{x}_{W_i}$, along with a small temporal map (i.e., SLAM features) $\mathbf{x}_{M_i}$ 
(see \cite{Geneva2020ICRA,Zhu2021ICRA}).
\begin{align} 
\mathbf{x}_{i,k} &= \begin{bmatrix} \mathbf{x}_{I_i}^\top 
& \mathbf{x}_{W_i}^\top & \mathbf{x}_{C_i}^\top & \mathbf{x}_{M_i}^\top
\end{bmatrix}^{\top}
\label{eq:state} \\
\mathbf{x}_{I_i} &= \begin{bmatrix} {}_G^{I_{i,k}} \bar{q}{}^{\top} & {}^G\mathbf{p}_{{I}_{i,k}}^{\top} & {}^G\mathbf{v}_{{I}_{i,k}}^{\top} & \mathbf{b}_{\omega_{i,k}}^{\top} & \mathbf{b}_{a_{i,k}}^{\top} 
\end{bmatrix}^{\top}  \\
\mathbf{x}_{W_i} &= 
\scalemath{0.95}{\begin{bmatrix}
{}^{C_i}t_{I_i} & {}^{C_i}_{I_i} \bar{q}{}^{\top} &  {}^{C_i}\mathbf{p}_{I_i}^{\top} & \boldsymbol{\zeta}_i^{\top}
\end{bmatrix}^{\top}}\\
\mathbf{x}_{C_i} &=
\scalemath{0.9}{\begin{bmatrix}
{}_G^{I_{i,k-1}} \bar{q}{}^{\top} &  {}^G\mathbf{p}_{I_{i,k-1}}^{\top} &
\cdots & 
{}_G^{I_{i,k-c}} \bar{q}{}^{\top} & {}^G\mathbf{p}_{I_{i,k-c}}^{\top}
\end{bmatrix}^{\top}}\\
\mathbf{x}_{M_i} &=
\scalemath{0.9}{\begin{bmatrix}
{}^G\mathbf{p}_{f1}^{\top} &
\cdots & 
{}^G\mathbf{p}_{fm}^{\top}
\end{bmatrix}^{\top}} 
\end{align}
where ${}_G^{I_{i,k}}\bar{q}$ is the unit quaternion parameterizing the rotation $\mathbf{C}({}_G^{I_{i,k}}\bar{q})={}_G^{I_{i,k}} \mathbf{R}$ from the global frame of reference $\{G\}$ to the IMU local frame $\{I_k\}$ at time $k$ for the $i$'th robot \cite{Trawny2005_Q_TR}, 
$\mathbf{b}_{\omega_{i,k}}$ and $\mathbf{b}_{a_{i,k}}$ are the gyroscope and accelerometer biases,
and ${}^G \mathbf{v}_{{I}_{i,k}}$ and ${}^G \mathbf{p}_{{I}_{i,k}}$ are the velocity and position of the IMU expressed in the global frame, respectively.
The clone state $\mathbf x_C$ contains $c$ historical IMU poses in a sliding window, while the temporal map state $\mathbf{x}_M$ has $m$ features.
{Each robot additionally calibrates its camera intrinsics $\boldsymbol{\zeta}_i$, camera-IMU extrinsics, and camera-IMU temporal offset ${}^{C_i}t_{I_i}$ \cite{Geneva2020ICRA}.}
Finally, given a group of $n$ robots, we have the following combined state and covariance matrix decomposition:
\begin{align}
    \mathbf{x}_k =
    \begin{bmatrix}
    \mathbf{x}_{1,k}^\top & \cdots & \mathbf{x}_{n,k}^\top
    \end{bmatrix}^\top \\
\scalemath{0.9}{
    \mathbf{P}_k =
    \begin{bmatrix}
    \mathbf{P}_{11_k} & \cdots & \mathbf{P}_{N1_k} \\
    \vdots & \ddots & \vdots \\
    \mathbf{P}_{1N_k} & \cdots & \mathbf{P}_{NN_k}
    \end{bmatrix} }
    \label{eq:cov}
\end{align}
Here we note that in the centralized formulation this is the state that we jointly estimate along with the cross-covariance terms, while in the distributed case each robot only estimates a sub-set of the total state and correlations between robots are dropped (e.g., robot $i$ only tracks $\mathbf{x}_{i,k}$ and $\mathbf{P}_{ii_k}$).

\subsection{Inertial Propagation} \label{sec:propagation}

The inertial state of the $i$'th robot $\mathbf{x}_{I_i}$ is propagated forward using its own IMU measurements of linear accelerations ($\mathbf a_{m_i}$) and angular velocities ($\bm\omega_{m_i}$) based on the following generic nonlinear IMU kinematics~\cite{Chatfield1997}:
\begin{align}
    \mathbf{x}_{i,k+1} = \mathbf{f}(\mathbf{x}_{i,k}, \mathbf a_{m_k}-\mathbf n_{a_k}, \bm\omega_{m_k}-\mathbf{n}_{\omega_k} )
    \label{eq:imu_dynamics}
\end{align}
where $\mathbf n_a$ and $\mathbf n_\omega$ are the zero-mean white Gaussian noise of the IMU measurements.
We linearize this nonlinear model at the current estimate for all robots, and can then propagate the state covariance matrix forward in time:
\begin{align}
\mathbf{P}_{k|k-1} &= \bm \Phi_{k-1}  \mathbf{P}_{k-1|k-1} \bm \Phi^{\top}_{k-1}+ \mathbf{Q}_{k-1} \label{eq:propcov} \\
\bm \Phi_{k-1} &= 
\mathbf{Diag}\left(\bm\Phi_{1,k-1}, \hdots, \bm\Phi_{N,k-1} \right)
\label{eq::big_phi} \\
\mathbf{Q}_{k-1} &= \mathbf{Diag}\left( \mathbf{Q}_{1,k-1}, \hdots, \mathbf{Q}_{N,k-1} \right) 
\end{align}
where $\bm\Phi_{i,k}$ and $\mathbf{Q}_{i,k}$ are respectively the system Jacobian and discrete noise covariance  for the $i$'th robot \cite{Mourikis2007ICRA}, and $\mathbf{Diag}(\cdots)$ creates a block diagonal matrix from the specified values.
In the distributed case, all states can be propagated independently since cross-covariance are not tracked.

\subsection{Camera Measurement Update}\label{sec:update}

A corner feature at time-step $k$ can be be written as the distortion of a perspective projection of a 3D point $^{C_{i,k}}\mathbf p_{f}$, expressed in the $i$'th robot's camera frame:
\begin{align} \label{eq:meas-model}
\mathbf z_{k} &= \mathbf{h}_{dist}(\mathbf z_{k,n}, \boldsymbol{\zeta}_i)  + \mathbf n_{f_k} \\
\mathbf z_{k,n}
&= \frac{1}{{}^{C_{i,k}}z_f} \begin{bmatrix} {}^{C_{i,k}}x_f \\ {}^{C_{i,k}}y_f \end{bmatrix} \\
{{}^{C_{i,k}}\mathbf p_{f}}  
&= {}^{C_i}_{I_i}\mathbf{R} ~{}^{I_{i,k}}_G\mathbf{R}
\left({^G\mathbf p_{f}} - {}^G\mathbf p_{I_{i,k}} \right) + {}^{C_i}\mathbf{p}_{I_i}
\label{eq:meas-eq}
\end{align}
where $\mathbf n_{f_k}$ is the zero-mean white Gaussian measurement noise with covariance $\mathbf R_{k}$, and $\mathbf{h}_{dist}(\cdot)$ is the camera distortion function which maps a normalized bearing $\mathbf z_{k,n}$ to the raw distorted image plane.
The linearization of this measurement model~\eqref{eq:meas-model} yields the following:
\begin{align} 
\mathbf r_{f_k}
&= \mathbf H_{k} \widetilde{\mathbf x}_{k} + \mathbf n_{f_k}  
= \mathbf H_{x_{i,k}} \widetilde{\mathbf x}_{I_{i}}
+ \mathbf H_{f_k} {^G\widetilde{\mathbf p}_{f}} + \mathbf n_{f_k} 
\label{eq:residual}
\end{align}
Once the measurement residual and Jacobian are computed the state and error covariance can be updated using the standard EKF update equations \cite{Maybeck1979}.

\section{Distributed Visual-Inertial CL}

As it is known that the standard EKF in the worst case has cubic computation complexity due to its covariance update,
a naive implementation of the multi-robot visual-inertial CL can become prohibitively expensive as the number of robots grow in size.
Note also that due to communication constraints, the robots might not be able to communicate with all the other robots or a common fusion center.
To address these issues, the key idea of our CL approach is to leverage CI~\cite{Julier2009HMDFTP} to reduce the estimation cost,
by only updating the state and error covariance of the current robot (i.e., robot $i$ only updates $\mathbf{x}_{i,k}$ and $\mathbf{P}_{ii_k}$) while ensuring consistency.

In particular, each robot  independently propagates its own state and updates with measurements that are only a function of its own state.
When updating with measurements of features  observed from multiple robots, 
CI is employed to consistently handle the unknown and untracked cross-covariance terms between the involved robots.
This means that robots need to communicate their state and covariance, along with visual feature information to the other robots.
Each robot tracks a set of visual features using KLT optical flow \cite{Kanade1981IJCAI},
and communicates its latest tracks and extracted ORB descriptors \cite{Rublee2011ICCV} to the other robots in communication range.
A robot then performs descriptor-based feature matching and loop-closure detection to find correspondences between its most recent features and other robots' feature tracks.
After tracking and matching,  feature tracks are categorized as follows:
\begin{itemize}
\item[(A)]  VIO features which have only been tracked for a short period of time.
\item[(B)]  Temporal SLAM features which have been tracked beyond the current sliding window.
\item[(C)]  Common VIO features which have been matched to features in another robot and tracked for only a short period of time.
\item[(D)]  Common SLAM features which have been matched to features in another robot. Note that this feature might be either a VIO  or SLAM feature in the other robot.
\end{itemize}
In the following, we present in detail how we update our state with these different feature variants.
Note that for the centralized case independent features update the full state and covariance since cross-covariances are tracked, 
while in the distributed case only the $i$'th robot state and covariance is updated thus allowing for computational savings.

\subsection{Independent VIO Feature: MSCKF Update} \label{sec:msckf_update}

For VIO features that have lost active track in the current window, we perform MSCKF update~\cite{Mourikis2007ICRA}.
In particular, we first triangulate these features for computing the feature Jacobians $\mathbf H_{f_k}$,
and then project $\mathbf{r}_{f_k}$ [see \eqref{eq:residual}] onto the left nullspace of $\mathbf{H}_{f_k}$ (i.e., $\mathbf{Q}_2^\top \mathbf H_{f_k} = \mathbf 0$) 
to yield the measurement noise independent of state:
\begin{align} \label{eq:msckf-null-trick}
    \mathbf{Q}_2^\top\mathbf{r}_{f_k} &= \mathbf{Q}_2^\top\mathbf{H}_{x_{i,k}}\tilde{\mathbf{x}}_{i,k} + \mathbf{Q}_2^\top\mathbf{H}_{f_k}{}^G\tilde{\mathbf{p}}_{f} + \mathbf{Q}_2^\top\mathbf{n}_{f_k} \\
   \Rightarrow~ \mathbf{r}'_{f_k} &= \mathbf{H}_x'\tilde{\mathbf{x}}_{k} + \mathbf{n}_{f_k}'
   \label{eq:feat_update_linearized}
\end{align}
where $\mathbf H_{x_{i,k}}$ is the stacked measurement Jacobians with respect to the navigation states in the current robot's  window.

\subsection{Independent SLAM Feature: FEJ-EKF Update}

SLAM features which a robot is able to reliably track longer then its sliding window in length, will be initialized into the SLAM map state vector $\mathbf{x}_{M_i}$.
{These features are directly updated using the linearized system \eqref{eq:residual} and will remain in the state until they have lost tracking.
To improve consistency, we employ First Estimate Jacobians (FEJ) \cite{Huang2008ISER,Huang2010IJRR} 
ensuring Jacobians are evaluated at the same linearization points to prevent spurious information gain.
}

\subsection{{Common VIO Feature: CI-EKF Update}} \label{sec:ci_update}

Consider we  find a feature which has been seen from multiple robots and want to use this information to update the state.
In the centralized case, we would   directly update our state with all available measurements~\eqref{eq:residual} through the standard EKF  since we track the cross-covariance  (e.g., $\mathbf{P}_{iN_k}$).
In the distributed case, a robot only tracks its own state and autocovariance to ensure computational efficiency and scalability with respect to the robot team size.
This presents two key challenges: (i) how to efficiently and consistently fuse multiple robots' autocovariances, and (ii) how to find the data association between different features,
which motivates us to leverage CI to fuse  estimates and covariances transmitted from other robots.

\subsubsection{CI-EKF Update}

Consider the $i$'th robot has a measurement which is a function of $L$ other robot states.
The linearized measurement model can be computed as:
\begin{align}
 \mathbf r_{f_k}
&= \mathbf H_{x_{i,k}} \widetilde{\mathbf x}_{i,k} +
\mathbf H_{x_{1..L,k}} \widetilde{\mathbf x}_{1..L,k}
+ \mathbf H_{f_k} {^G\widetilde{\mathbf p}_{f}}
+ \mathbf n_{f_k}
\label{eq:residual_multi}
\end{align}
where $\mathbf H_{x_{i,k}}$ is the Jacobian in respect to the $i$'th robot state using the $k$'th estimates, and $\mathbf H_{x_{1..L,k}}$ is the stacked Jacobian with respect to all other robots the measurement is a function of.
To guarantee consistency when updating with this measurement, we adopt the CI-EKF update~\cite{Julier2009HMDFTP} to construct a prior covariance such that:
\begin{align}
\mathbf{Diag}
\left(
\sfrac{1}{\omega_i}\mathbf{P}_{ii_k},
\sfrac{1}{\omega_1}\mathbf{P}_{11_k}, \cdots, \sfrac{1}{\omega_L}\mathbf{P}_{LL_k}
\right) 
\geq \mathbf{P}_k
\label{eq:ci}
\end{align} 
where the left side is the CI covariance with zero off-diagonal elements and the right hand side is the unknown true covariance of the state with cross-covariances [see \eqref{eq:cov}].
The weights $\omega_l>0$ and $\sum_{l}\omega_{l} = 1$, for $l \in \{i,1..L\}$, can be found optimally \cite{Julier2009HMDFTP}.
Substituting \eqref{eq:ci} into the standard EKF equations and only selecting the portion that updates the current robot's state (say robot $i$)  yields:
\begin{align}
\delta{\mathbf{x}_{i,k}} &= \frac{1}{\omega_i}\mathbf{P}_{ii,k|k-1}\mathbf{H}_{x_{i,k}}^\top\mathbf{S}_{k}^{-1}{\mathbf{r}}_{f_k}' \label{eq:cistate}\\
\mathbf{P}_{ii,k|k} &= \frac{1}{\omega_i}\mathbf{P}_{ii,k|k-1}  
\!-\! \frac{1}{\omega_i^2}\mathbf{P}_{ii,k|k-1} \! \mathbf{H}_{x_{i,k}}^\top \! \mathbf{S}_{k}^{-1}\! \mathbf{H}_{x_{i,k}}\mathbf{P}_{ii,k|k-1} \label{eq:cicov}\\
\mathbf{S}_k &= \sum_{o \in \{i,1..L\}}\frac{1}{\omega_o}
\mathbf{H}_{x_{o,k}}\mathbf{P}_{oo,k|k-1}\mathbf{H}_{x_{o,k}}^\top
+ \mathbf{R}_{f_k} \label{eq:cicovs_s}
\end{align}
where $\delta{\mathbf{x}_{i,k}}$ is the correction to the state estimate $\hat{\mathbf x}_{i,k}$.

\subsubsection{{Efficient Nullspace Projection}} \label{sec:ci_msckf_update}

To process common features which are short in length, we leverage the similar logic as in Sec. \ref{sec:msckf_update}.
For example,  we have multiple measurements from two different robots and wish to update our state:
\begin{align}
\setlength\arraycolsep{2pt}
\begin{bmatrix}
\mathbf r_{f_{i,k}} \\
\mathbf r_{f_{2,k}} 
\end{bmatrix}
=
\begin{bmatrix}
\mathbf H_{x_{i,k}} & \mathbf{0} & \mathbf H_{f_{i,k}} \\
\mathbf{0} & \mathbf H_{x_{2,k}} & \mathbf H_{f_{2,k}}
\end{bmatrix}
\begin{bmatrix}
\widetilde{\mathbf x}_{I_{i}} \\
\widetilde{\mathbf x}_{I_{2}} \\
{^G\widetilde{\mathbf p}_{f}}
\end{bmatrix}
+
\begin{bmatrix}
\mathbf n_{f_{i,k}} \\
\mathbf n_{f_{2,k}}
\end{bmatrix} \label{eq:original_ci_msckf}
\end{align}
We can then project both equations onto their left range and nullspace (e.g., $\mathbf{H}_{f_{i,k}}=[\mathbf{Q}_{i,1}~\mathbf{Q}_{i,2}][\mathbf{U}_{i} ~\mathbf{0}]^\top$):
\begin{equation}
\setlength\arraycolsep{2pt}
\begin{bmatrix}
\mathbf r_{f_{i,k}}^1 \\
\mathbf r_{f_{i,k}}^2 \\
\mathbf r_{f_{2,k}}^1 \\
\mathbf r_{f_{2,k}}^2 
\end{bmatrix}
=
\begin{bmatrix}
\mathbf{Q}_{i,1}^\top \mathbf  H_{x_{i,k}} & \mathbf{0} & \mathbf{U}_{i} \\
\mathbf{Q}_{i,2}^\top \mathbf H_{x_{i,k}} & \mathbf{0} & \mathbf{0} \\
\mathbf{0} & \mathbf{Q}_{2,1}^\top \mathbf H_{x_{2,k}} & \mathbf{U}_{2} \\
\mathbf{0} & \mathbf{Q}_{2,2}^\top \mathbf H_{x_{2,k}} & \mathbf{0}
\end{bmatrix}
\begin{bmatrix}
\widetilde{\mathbf x}_{I_{i}} \\
\widetilde{\mathbf x}_{I_{2}} \\
{^G\widetilde{\mathbf p}_{f}}
\end{bmatrix}
+
\begin{bmatrix}
\mathbf n_{f_{i,k}}^1 \\
\mathbf n_{f_{i,k}}^2 \\
\mathbf n_{f_{2,k}}^1 \\
\mathbf n_{f_{2,k}}^2
\end{bmatrix} \notag
\end{equation}
where we have defined that $\mathbf r_{f_{i,k}}^1=\mathbf{Q}_{i,1}^\top \mathbf r_{f_{i,k}}$
and $\mathbf n_{f_{i,k}}^1=\mathbf{Q}_{i,1}^\top \mathbf n_{f_{i,k}}$.
Note that the last row is no longer dependent on the current robot's state, ${\mathbf x}_{I_{i}}$, and thus,
this can be discarded since it will not update the state or covariance due to the lack of tracked cross-covariances.
This directly reduces the number of measurements involved during update and makes the computation of $\mathbf{S}_{k}^{-1}$  substantially cheaper [see \eqref{eq:cicovs_s}].
We then have the following linear systems:
\begin{align}
\setlength\arraycolsep{2pt}
\begin{bmatrix}
\mathbf r_{f_{i,k}}^1 \\
\mathbf r_{f_{2,k}}^1 \\
\end{bmatrix} &=
\begin{bmatrix}
\mathbf{Q}_{i,1}^\top \mathbf  H_{x_{i,k}} & \mathbf{0} & \mathbf{U}_{i} \\
\mathbf{0} & \mathbf{Q}_{2,1}^\top \mathbf H_{x_{2,k}} & \mathbf{U}_{2} \\
\end{bmatrix}
\begin{bmatrix}
\widetilde{\mathbf x}_{I_{i}} \\
\widetilde{\mathbf x}_{I_{2}} \\
{^G\widetilde{\mathbf p}_{f}}
\end{bmatrix}
+\begin{bmatrix}
\mathbf n_{f_{i,k}}^1 \\
\mathbf n_{f_{2,k}}^1 \\
\end{bmatrix} \notag \\
\mathbf r_{f_{i,k}}^2 &=
\mathbf{Q}_{i,2}^\top \mathbf H_{x_{i,k}} \widetilde{\mathbf x}_{I_{i}}
+\mathbf n_{f_{i,k}}^2 \label{eq:independent_ci_msckf}
\end{align}
A second nullspace projection onto the left nullspace of $\mathbf{H}_f=[\mathbf{U}_{i} ~\mathbf{U}_{2}]^\top$ is performed to create a linear system which is only a function of the ${\mathbf x}_{I_{i}}$ and ${\mathbf x}_{I_{2}}$ states.
The CI-EKF update  [see  \eqref{eq:cistate} and \eqref{eq:cicov}] is then used to update the state ${\mathbf x}_{I_{i}}$.
The second equation [see \eqref{eq:independent_ci_msckf}] can update the current robot state without CI through the standard EKF equations since it is only a function of the current robot state.
This update contains the same information as in the case that we performed a ``large'' nullspace projection using the full feature Jacobians in \eqref{eq:original_ci_msckf}, 
but results in a much smaller measurement size since we can drop measurement residuals which are not a function of the $i$'th robot's state.

\subsection{Common SLAM Feature: CI-EKF Update} \label{sec:ci_update_slam}

There are two different cases for temporal SLAM features:
(i) a SLAM feature in the current robot state matches to a feature that is not a SLAM feature in another robot, and
(ii) a SLAM feature matches to another robot's SLAM feature.
For example as in Fig. \ref{fig:diagram}, 
in the first case we collect the measurements from the other robot ($\mathbf{z}_{1..N}$) and directly apply \eqref{eq:residual_multi} and update both the current robot's poses and its estimate of the SLAM feature.
In the second case, we can either follow this same logic (i.e., grab the measurements from the other robot and update current robot's estimate) 
or we can leverage the knowledge that the 3D position of these two features should be equal.
This SLAM feature constraint model is similar to the one introduced in \cite{Guo2018TRO} for cooperative mapping.
Consider we have the following two robots:
\begin{align}
    \mathbf{x}_{i,k} &=
    \begin{bmatrix}
    \mathbf{x}_{I_i}^\top & \mathbf{x}_{W_i}^\top & \mathbf{x}_{C_i}^\top &
    {}^G\mathbf{p}_{fa}^\top
    \end{bmatrix}^\top \\
    \mathbf{x}_{2,k} &=
    \begin{bmatrix} 
    \mathbf{x}_{I_2}^\top & \mathbf{x}_{W_2}^\top & \mathbf{x}_{C_2}^\top &
    {}^G\mathbf{p}_{fb}^\top
    \end{bmatrix}^\top
\end{align}
If we have matched feature ${}^G\mathbf{p}_{fa}$ in the current $i$'th robot to the ${}^G\mathbf{p}_{fb}$ in the other robot, 
then we can construct the following feature constraint (see Fig. \ref{fig:diagram}):
\begin{align} 
{}^G\mathbf{p}_{fa} - {}^G\mathbf{p}_{fb} &= \mathbf{0} 
\Rightarrow~ \mathbf{r}_{c}\left(\mathbf{x}_{i,k},\mathbf{x}_{2,k}\right) = \mathbf{0}
\end{align}
which can be linearized to yield:
\begin{align}
&\mathbf{r}_c\left(\hat{\mathbf{x}}_{i,k},\hat{\mathbf{x}}_{2,k}\right)
+ \mathbf H_{fa} {^G\widetilde{\mathbf p}_{fa}}
+ \mathbf H_{fb} {^G\widetilde{\mathbf p}_{fb}} 
\approx \mathbf{0} \\
\Rightarrow &
\mathbf{0} - \mathbf{r}_c\left(\hat{\mathbf{x}}_{i,k},\hat{\mathbf{x}}_{2,k}\right)
\approx \mathbf H_{fa} {^G\widetilde{\mathbf p}_{fa}}
+ \mathbf H_{fb} {^G\widetilde{\mathbf p}_{fb}} \label{eq:constraint_res}
\end{align}
This linearized system can then update the $i$'th robot state estimate using the CI-EKF update  [see \eqref{eq:cistate} and \eqref{eq:cicov}].
Note that this is a very efficient update, as it is only a function of the two estimated feature positions.

\subsection{Historical Features: CI-EKF Update}

\begin{figure}\centering
\includegraphics[width=0.9\columnwidth]{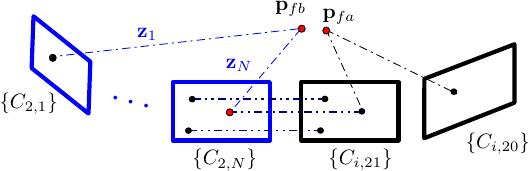}
\caption{
Illustration of the keyframe-aided 2D-to-2D matching for data association. 
Assuming robot $i$'s 21st frame $\{C_{i,21}\}$ matches to the 2nd robot's $N$'th frame $\{C_{2,N}\}$.
We are able to find all feature correspondences between the features the robot's observer, namely $\mathbf{z}_{1..N}$.
}
\label{fig:diagram}
\vspace{-2em}
\end{figure}

We now explain how to leverage loop-closure constraints to previous robot states.
First, to find the feature correspondences between robots, as in \cite{Geneva2019ICRA,Geneva2019CVPR}, each robot create DBoW2 \cite{Galvez2012TRO} databases for all other robots.
When a robot receives feature tracks and descriptors from other robots they are appended to their corresponding DBoW2 database.
The current image can then be queried against the other robots' databases to see if any other robots are or have been at the current location.
If a loop-closure is detected and verified using a fundamental matrix geometric check, then we assume that we have detected that another robot has been at our current location.
After matching  descriptors, we  know the correspondences between a feature in the current robot, and that of the features in the other robot (see Fig. \ref{fig:diagram}).
We can then grab the history of measurements and formulate a common feature update.

To incorporate these measurements from historical states, each robot records the measurement and previous states received from the other agents.\footnote{{In the future we plan to investigate the latency introduced due to communication constraints, but historical matching ensures that the robot will leverage \textit{all} available information at the current time including delayed information recently communicated.}}
Outside of the most recent sliding window, these historical states can provide loop-closure information if we are able to generate measurement constraints to them.
Specifically we store the following historical states and covariances in addition to their most recent states published:
\begin{align}
    \mathbf{x}_i &=
    \left\{
    \mathbf{x}_{i,0},\cdots,\mathbf{x}_{i,k-1}
    \right\}, ~
    \mathbf{P}_i =
    \left\{
    \mathbf{P}_{ii_0},\cdots,\mathbf{P}_{ii_{k-1}}
    \right\}
\end{align}
Since each one of these historical states contain a sliding window of poses and SLAM features, we only store non-overlapping sliding windows. 
To accelerate lookup we only store historical descriptor information at a fixed rate (normally 1Hz) since recent frames in the same sliding window contain redundant loop-closure information.
More ideal heuristic could be leveraged here to increase match rates.
Once loop-closure is detected, we know old historical feature correspondences which we can then use to retrieve measurements and update our current robot state.
This update is identical to the CI-EKF update as in Sec. \ref{sec:ci_update}--\ref{sec:ci_update_slam},
which only needs to involve the historical windows that contain the historical measurements, and thus is efficient since historical states are not updated.

\color{black}
 
\begin{table}\centering
\caption{
Simulation parameters and prior standard deviations that perturbations of measurements and initial states were drawn from.
}
\label{tab:sim_params}
\begin{adjustbox}{width=\columnwidth,center}
\begin{tabular}{ccccc} \toprule
\textbf{Parameter} & \textbf{Value} & \textbf{Parameter} & \textbf{Value} \\ \midrule
Gyro. White Noise & 1.6968e-04 & Gyro. Rand. Walk & 1.9393e-05 \\
Accel. White Noise & 2.0000e-3 & Accel. Rand. Walk & 3.0000e-3 \\
Pixel Proj. (px) & 1 & Robot Num. & 3 \\
IMU Freq. (hz) & 400 &  Cam Freq. (hz) & 10 \\
AR Avg. Feats & 25 & AR Num. SLAM & 3  \\
ETH Avg. Feats & 50 & ETH Num. SLAM & 5  \\
Num. Clones & 11 & Feat. Rep. & GLOBAL \\\bottomrule
\end{tabular}
\end{adjustbox}
\vspace{-1em}
\end{table}
\begin{figure}
\centering
\includegraphics[width=.45\columnwidth]{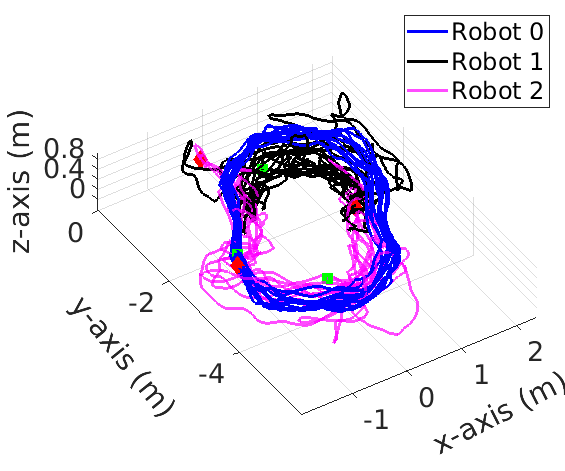}
\includegraphics[width=.45\columnwidth]{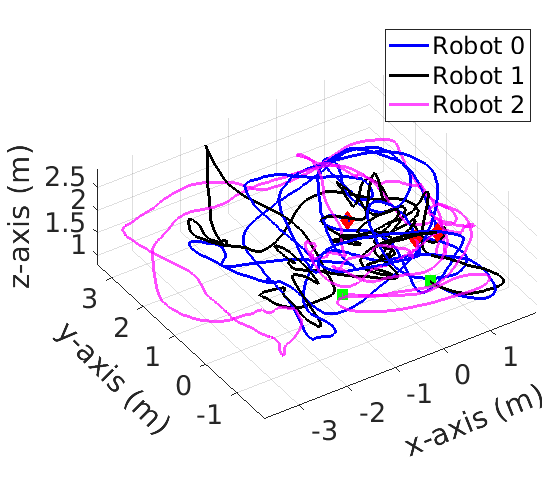}
\caption{
Simulated trajectories, axes are in units of meters.
General hand-held AR dataset (left) are 147, 93, and 100 meters long, while ETH EuRoC MAV Vicon room datasets (right) are 70, 58, and 59 meters long for each robot.
Green square denotes the start and red diamond denotes the end.
}
\label{fig:simtraj}
\vspace*{-2em}
\end{figure}

\section{Simulation Results}

To validate the proposed method, we have simulated two  realistic scenarios both   with three robots (see Fig. \ref{fig:simtraj}).
The first is a hand-held mobile AR dataset which has a series of users look and move around a central table,
while the second is a series of trajectories from the ETH EuRoC MAV dataset \cite{Burri2016IJRR}.
We employ the OpenVINS simulator \cite{Geneva2019ICRA} to generate realistic visual-bearing and inertial measurements from these supplied trajectories.
On average each robot is able to find common features on, respectively, $79.0\%$ and $83.5\%$ ($43.7\%$ and $62.7\%$) of the frames without or with loop-closure in AR datasets (ETH dataset).
{This clearly shows that advantage of historical loop-closure on datasets which have limited temporal view overlaps between robots.}
Simulation parameters used are documented in Tab. \ref{tab:sim_params}.
We fix the weight of other robots' covariance in the CI-EKF update as $\omega_o=0.001$.
While for the constraint measurement update presented in Sec. \ref{sec:ci_update_slam}, we use the value $\omega_o=0.005$ and a synthetic measurement noise of 2cm.
Note that while these weights can be found by minimizing the trace or determinant of $\mathbf{P}_{ii,k|k}$ \cite{Julier2009HMDFTP},  
we have empirically found that using fixed weights still ensures consistent performance.
For fair and thorough comparison, we define the following variations of the centralized and proposed distributed CL estimators:
\begin{itemize}[leftmargin=!,labelindent=-5pt,itemindent=-10pt]
    \item[] \textbf{indp} -- No common features are found between robots and all measurements are processed as independent features which only relate to the current robot.
    \item[] \textbf{indp-slam} -- Same as \textit{indp}, but temporal SLAM features are included in each robot to show the relative improvement.
    \item[] \textbf{ce-cmsckf} -- The centralized estimator using the common VIO features over the sliding window.
    \item[] \textbf{ce-cmsckf-cslam} -- The centralized estimator using the common VIO and  SLAM features over the sliding window. 
    \item[] \textbf{dc-cmsckf} \cite{Zhu2021ICRA} -- The distributed estimator using the common VIO features over the sliding window.
    \item[] \textbf{dc-cmsckf-cslam} -- The distributed estimator using the common VIO and  SLAM features over the sliding window without enforcing the same feature constraint. For example, even if a common SLAM feature is a SLAM feature in another robot's state, we grab the measurements from the other robot and update as the first case in Sec. \ref{sec:ci_update_slam}. 
    \item[] \textbf{dc-full-window} -- The distributed estimator using the common VIO and  SLAM features over the sliding window with enforcing the same feature constraint.
    \item[] \textbf{dc-full-history} -- The distributed estimator using both the common VIO and  SLAM features over the sliding window and from historical matching. 
\end{itemize} 
Note that the observed independent VIO features and SLAM features are used in all these estimators.
To ensure a fair comparison, the same parameters reported in Tab. \ref{tab:sim_params} are used for all algorithms and for all robots.

\subsection{Accuracy and Consistency Evaluation}
\begin{table}[t]
\centering
\caption{
ATE on simulated AR datasets in degrees / meters for each algorithm variation. Green denotes the best, while blue is second best. 
}
\begin{adjustbox}{width=0.95\columnwidth,center}
\begin{tabular}{rcccccc} \toprule 
\textbf{Algorithm} & \textbf{Robot 0} & \textbf{Robot 1} & \textbf{Robot 2} & \textbf{Average} \\\midrule
indp & 1.957 / 0.072 & 0.811 / 0.041 & 0.742 / 0.039 & 1.170 / 0.051 \\
indp-slam & 1.396 / 0.046 & 0.602 / 0.029 & 0.557 / 0.022 & 0.852 / 0.032 \\\midrule
ce-cmsckf & 0.364 / 0.017 & 0.323 / 0.015 & 0.355 / 0.015 & 0.347 / 0.016  \\
ce-cmsckf-cslam & \gb{0.232} / \gb{0.011} & \gb{0.228} / \gb{0.011} & \gb{0.220} / \gb{0.010} & \gb{0.227} / \gb{0.011} \\
\midrule
dc-cmsckf & 0.759 / 0.029 & 0.540 / 0.025 & 0.553 / 0.020 & 0.617 / 0.025 \\
dc-cmsckf-cslam & 0.643 / 0.025 & 0.496 / 0.022 & 0.478 / 0.017 & 0.539 / 0.022 \\
dc-full-window & 0.644 / 0.024 & 0.547 / 0.022 & 0.480 / 0.017 & 0.557 / 0.021 \\
dc-full-history & \gs{0.356} / \gs{0.017} & \gs{0.299} / \gs{0.014} & \gs{0.319} / \gs{0.013} & \gs{0.325} / \gs{0.014} \\
\bottomrule\\
\end{tabular}
\label{tab:sim_ate_ar}
\end{adjustbox}
\vspace*{-3em}
\end{table}
\begin{table}[t]
\centering
\caption{
ATE on simulated ETH datasets in degrees / meters for each algorithm variation. Green denotes the best, while blue is second best.
}
\begin{adjustbox}{width=0.95\columnwidth,center}
\begin{tabular}{rcccccc} \toprule 
\textbf{Algorithm} & \textbf{Robot 0} & \textbf{Robot 1} & \textbf{Robot 2} & \textbf{Average} \\\midrule
indp & 0.569 / 0.088 & 0.578 / 0.092 & 0.560 / 0.093 & 0.569 / 0.091 \\
indp-slam & 0.371 / 0.070 & 0.406 / 0.069 & 0.444 / 0.075 & 0.407 / 0.071 \\\midrule
ce-cmsckf & 0.221 / 0.052 & 0.221 / 0.049 & 0.221 / 0.051 & 0.221 / 0.050 \\
ce-cmsckf-cslam & \gb{0.151} / \gs{0.042} & \gb{0.143} / \gs{0.038} & \gb{0.144} / \gs{0.040} & \gb{0.146} / \gs{0.040} \\\midrule
dc-cmsckf & 0.329 / 0.064 & 0.342 / 0.061 & 0.319 / 0.062 & 0.330 / 0.062 \\
dc-cmsckf-cslam & 0.298 / 0.054 & 0.325 / 0.050 & 0.290 / 0.052 & 0.304 / 0.052 \\
dc-full-window & 0.285 / 0.052 & 0.287 / 0.047 & 0.268 / 0.047 & 0.280 / 0.049 \\dc-full-history & \gs{0.211} / \gb{0.029} & \gs{0.207} / \gb{0.031} & \gs{0.218} / \gb{0.030} & \gs{0.212} / \gb{0.030} \\
\bottomrule
\end{tabular}
\label{tab:sim_ate_eth}
\end{adjustbox}
\end{table}
\begin{figure}[t]
\centering
\includegraphics[width=.46\columnwidth]{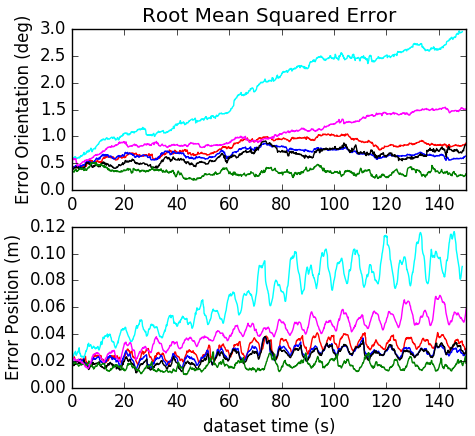}
\vspace*{0.2em}
\includegraphics[width=.46\columnwidth]{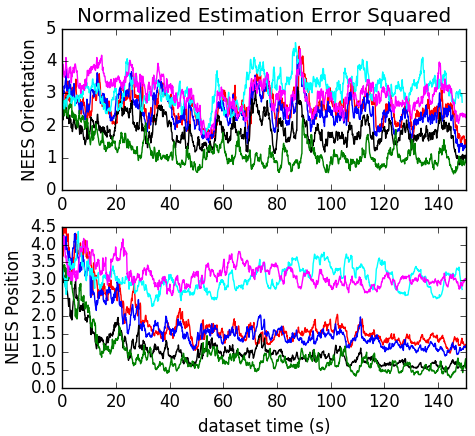}
\includegraphics[width=.46\columnwidth]{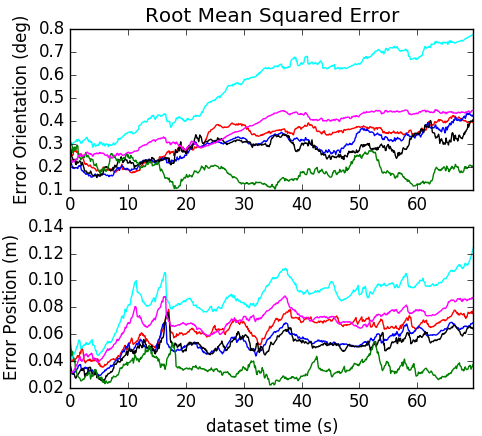}
\includegraphics[width=.46\columnwidth]{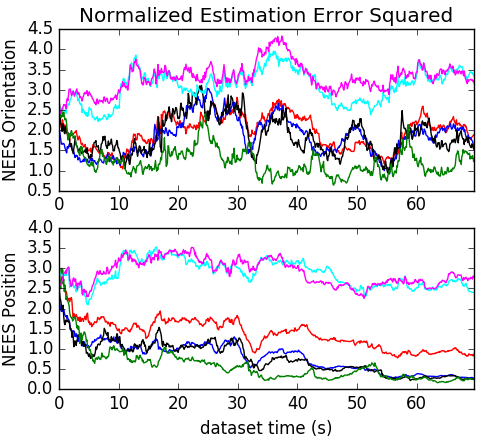}
\caption{ Robot 0's average RMSE (left) and NEES (right) results in the simulated AR (top) and ETH datasets (bottom). Cyan represents indp, magenta represents indp-slam, red represents dc-msckf, blue represents dc-cmsckf-cslam, green represents dc-full-window and green represent dc-full-history.
Please refer to the color figure.
}
\label{fig:sim_rmse_robot0}
\vspace*{-2em}
\end{figure}

We performed 20 Monte Carlo simulations on each dataset.
The average Absolute Trajectory Error (ATE) \cite{Zhang2018IROS} can be found in Tab. \ref{tab:sim_ate_ar} and \ref{tab:sim_ate_eth}.
It is clear from the top two rows that the additional SLAM features improve \texttt{indp}.
In the cooperative case, when using the common VIO features, both \texttt{ce-msckf} and \texttt{dc-msckf} outperform the \texttt{indp-slam}, and when including common SLAM features, the accuracy is further improved.
It is worth noting that the efficient \texttt{dc-full-window} with feature constraint has close accuracy to its counterpart \texttt{dc-cmsckf-cslam}.
Moreover, when including the historical common features, the distributed estimator becomes more accurate as expected.
Interestingly, with only the common features over the sliding window, the \texttt{ce-cmsckf-cslam} can achieve the best performance on the AR dataset even without loop-closure.
This is likely due to the fact that over the whole dataset all robots look in the same general location thus negating any benefit of loop-closure detection.
As show in Tab. \ref{tab:sim_ate_eth} and in the following real-world experiments, when robots do not have many overlapping views, the historical information plays an important role.

We additionally show the average {Root Mean Square Error (RMSE) \cite{Zhang2018IROS} and Normalized Estimation Error Squared (NEES) \cite{Bar2004}} of the distributed algorithms for Robot $0$ in Fig. \ref{fig:sim_rmse_robot0}.
The results for the other two robots are similar and are omitted here for space.
The \texttt{indp} has the largest drift that can be reduced as shown by \texttt{indp-slam} and leveraging common features.
The \texttt{dc-cmsckf-cslam} and \texttt{dc-full-window} have almost the same performance while the \texttt{dc-full-history} achieves the best accuracy.
It is clear that all the distributed algorithms are conservative in nature (NEES is smaller than three) and have smaller NEES than the centralized ones.

\subsection{Timing Analysis}

\begin{figure}\centering
\includegraphics[width=0.90\columnwidth]{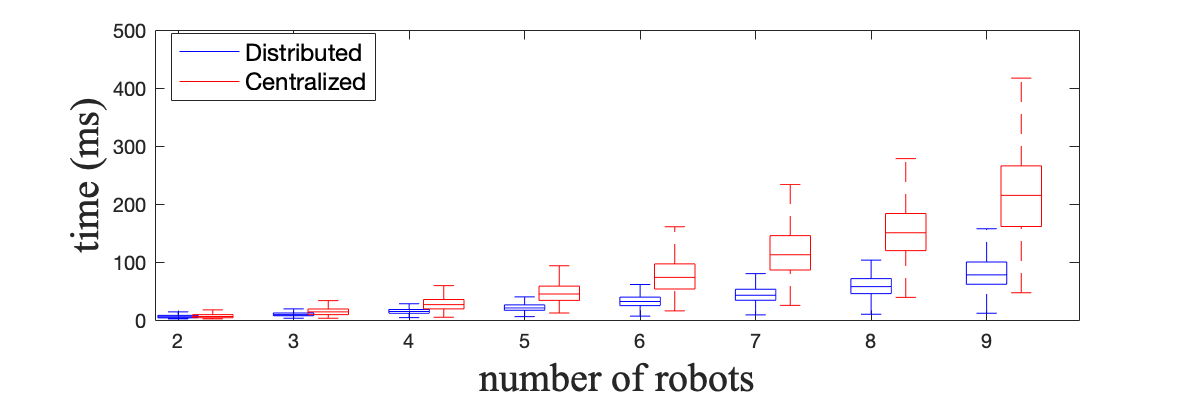}
\caption{
Sequential propagation and update time (ms). Note that while decentralized can update in parallel, here we report its sequential timings.
}
\label{fig:sim_timing_multirobot}
\vspace{-0.5em}
\end{figure}
\begin{table}[t]
\centering
\caption{
{Timing for AR dataset. Millisecond mean and deviation.}
}
\label{tab:sim_timing_msckf_nullspace}
\begin{tabular}{rccccc} \toprule
\textbf{Algorithm} & \textbf{Proposed} & \textbf{Combined}\\\midrule
MSCKF update (window) & 1.20  $\pm$ 0.94    & 2.88  $\pm$ 3.90 \\
MSCKF update (hist)   & 4.11  $\pm$ 5.52    & 22.75   $\pm$ 159.65 \\
\end{tabular}
\label{tab:sim_timing_constraint}
\begin{tabular}{rccccc} \toprule 
\textbf{Algorithm} & \textbf{Constraint} & \textbf{No Constraint} \\\midrule
SLAM update (window) & 0.10  $\pm$ 0.03    & 0.15   $\pm$ 0.16 \\
SLAM update (hist)   & 0.17   $\pm$ 0.06    & 27.11  $\pm$ 160.95 \\
\bottomrule
\end{tabular}
\vspace*{-2em}
\end{table}

\subsubsection{Multiple Robots}
We now investigate the computational efficiency of the proposed work in comparison to the centralized estimator using only common features over the sliding window. 
We compare the timing results of \texttt{dc-full-window} and \texttt{ce-cmsckf-cslam} while processing the same amount of measurements.
We first investigate the performance as more robots are added to show the efficiency gains from the distributed formulation.
The results in Fig. \ref{fig:sim_timing_multirobot} show that as more robots are added, the centralized estimator quickly becomes {computationally expensive}
while the distributed one is able to remain efficient since each robot only needs to propagate and update its own state and auto-covariance. Additionally, if one robot does not find common features in a given frame, the robot can update the estimator independently in the distributed case.
On the contrary, the centralized estimator needs to collected all data, propagate, and update the whole state even if there are no common features.
The distributed algorithm does have a slight increase in cost, which is due to the increase of common measurements from the additional robots.

\subsubsection{Common VIO Features}
We next investigate the efficiency of the common VIO feature nullspace projection and subsequent CI-EKF update introduced in Sec. \ref{sec:ci_msckf_update}.
We report the update time for \texttt{dc-full-window} (window) and \texttt{dc-full-history} (history) without common SLAM features.
The results presented in Tab. \ref{tab:sim_timing_msckf_nullspace} show that if we use the proposed method to first perform nullspace projection and separate each robot's systems into two systems (Proposed) we are able to outperform the naive way of performing nullspace projection on a ``stacked'' Jacobian containing all robot feature Jacobians (Combined).
It is clear that in both algorithms, the proposed method is able to have less computational cost, especially in the historical case due to the proposed system reducing the number of measurements in the update. 
We also note that there is a high level of variance in the historical case due to loop-closure introducing large amounts of measurements in short intervals.

\subsubsection{SLAM Constraint Update}

Now we investigate the efficiency of the common SLAM feature update introduced in Sec. \ref{sec:ci_update_slam}.
Only common SLAM features that can be matched to another robot's SLAM feature are used to ensure that both variants have the same number of measurements in the update.
When we match features in the current window, the constraint update (Constraint) is slight more efficient than the naive way of grabbing all the measurements from the other robots (No Constraint) since all robots only have the most recent measurements (in most cases just one).
During historical SLAM matching, by definition SLAM features are long feature tracks, and thus many measurements and clones states are associated with a historical SLAM feature.
This means that after loop-closure in the naive case (No Constraint) we will process all measurements ever recorded for a SLAM feature which can easily reach many sliding windows in length.
If instead we use the constraint update, only the two feature positions are involved, thus the update is extremely efficient in nature (bottom Tab. \ref{tab:sim_timing_constraint}).

\begin{figure}[t]
\centering
\includegraphics[width=.49\columnwidth]{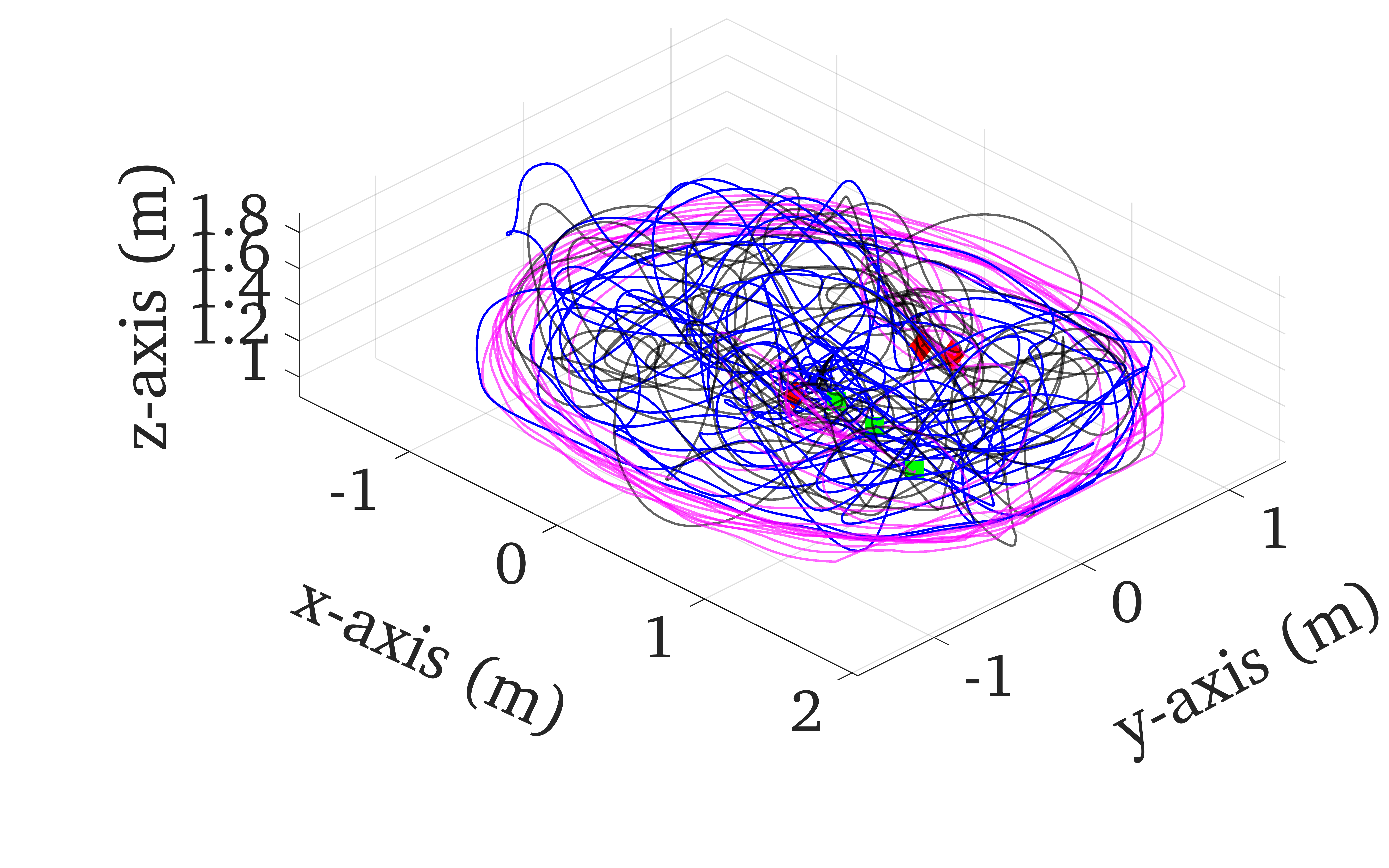}
\includegraphics[width=.49\columnwidth]{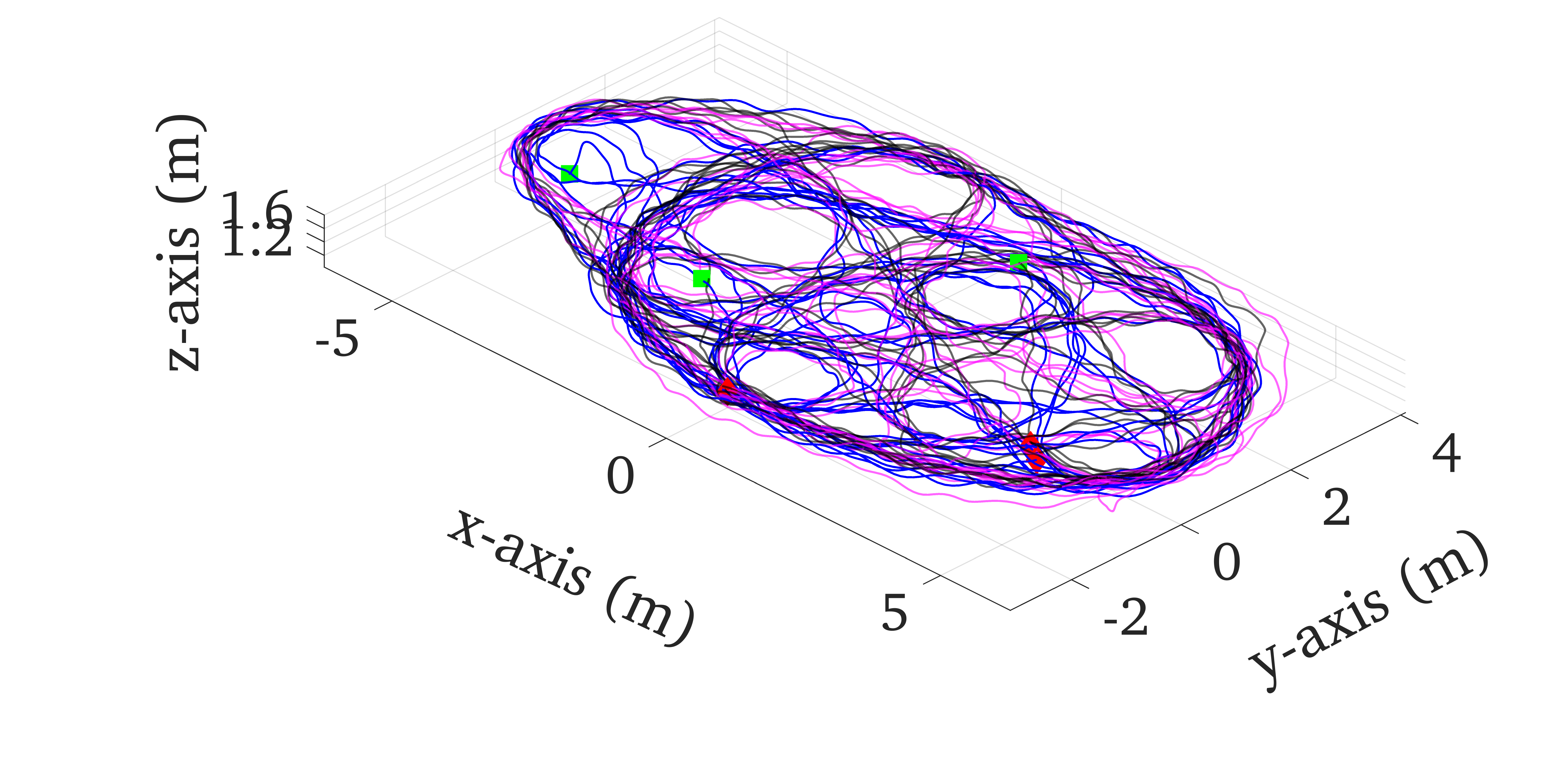}
\caption{
TUM-VI groundtruth (left) and Vicon room groundtruth trajectories (right)
TUM-VI trajectories are 146, 131, and 134 meters long, while the Vicon room datasets are 507, 509, and 501 meters long.
}
\label{fig:exptraj}
\vspace{-0.8em}
\end{figure}
\begin{table}[t]
\centering 
\caption{
Relative pose error (RPE) on TUM-VI datasets in degrees / meters averaged over all robots for the dataset.
}
\begin{adjustbox}{width=0.99\columnwidth,center}
\begin{tabular}{rcccccc} \toprule 
\textbf{Algorithm} & \textbf{40m} & \textbf{60m} & \textbf{80m} & \textbf{100m} & \textbf{120m} \\\midrule
indp-slam & 1.818 / 0.093 & 2.833 / 0.126 & 2.604 / 0.154 & 2.774 / 0.185 & 2.716 / 0.215 \\
\midrule
ce-cmsckf & \gs{1.358} / 0.071 & \gs{1.321} / 0.091 & 1.357 / 0.108 & 0.843 / 0.128 & {0.932} / 0.140 \\
ce-cmsckf-cslam & 1.758 / \gs{0.069} & 1.350 /\gs{0.079} &\gb{1.027} / \gs{0.100} & \gb{0.718} / {0.119} & {0.938} / \gs{0.130} \\
\midrule
dc-cmsckf & 1.662 / 0.075 & 2.005 / 0.104 & 1.605 / 0.129 & 1.142 / 0.141 & 1.531 / 0.170 \\
dc-cmsckf-cslam & 1.800 / 0.080 & 2.642 / 0.093 & 2.233 / 0.106 & 1.544 / \gs{0.114} & 0.934 / 0.157 \\
dc-full-window & 1.768 / 0.075 & 2.218 / 0.091 & 1.788 / 0.109 & 1.257 / 0.123 & \gs{0.854} / 0.159 \\
dc-full-history & \gb{1.213} / \gb{0.067} & \gb{1.232} / \gb{0.061} & \gs{1.029} / \gb{0.065} & \gs{1.004} / \gb{0.068} & \gb{0.784} / \gb{0.072} \\
\bottomrule
\end{tabular}
\label{tab:exp_tum}
\end{adjustbox}
\vspace*{0.5em}
\centering 
\caption{
Relative pose error (RPE) on Vicon room dataset in degrees / meters averaged over all robots.
}
\begin{adjustbox}{width=0.99\columnwidth,center}
\begin{tabular}{rcccccc} \toprule 
\textbf{Algorithm} & \textbf{80m} & \textbf{100m} & \textbf{200m} & \textbf{300m} & \textbf{420m} \\\midrule
indp-slam & \gs{2.022} / \gs{0.276} & 2.416 / 0.334 & 3.872 / 0.613 & 5.222 / 0.870 & 8.045 / 1.189 \\
\midrule
ce-cmsckf-cslam & 2.180 / 0.288 & 2.603 / 0.333 & \gs{2.771} / \gs{0.548} & \gs{3.050} / \gs{0.770} & \gs{3.557} / \gs{1.044} \\\midrule
dc-full-window & 2.197 / 0.281 & \gs{2.340} / \gs{0.332} & 3.322 / 0.580 & 3.670 / 0.804 & 5.977 / 1.102 \\
dc-full-history & \gb{1.271} / \gb{0.145} & \gb{1.307} / \gb{0.151} & \gb{1.346} / \gb{0.158} & \gb{1.267} / \gb{0.157} & \gb{1.343} / \gb{0.160} \\
\bottomrule
\end{tabular}
\label{tab:exp_cyclops}
\end{adjustbox}
\vspace*{-2em}
\end{table}

\section{Experimental Results}

We have also evaluated the proposed distributed CL estimators on the TUM-VI dataset \cite{Schubert2018IROS} and a hand collected 10 minute long Vicon room dataset (see Fig. \ref{fig:exptraj}).\footnote{A video demo \href{https://youtu.be/boHBcVoMKk8}{https://youtu.be/boHBcVoMKk8}}
Both datasets provide monochrome stereo images at 20Hz and IMU readings at 200Hz.
We only leverage the left camera and initialize all robots based on the groundtruth orientation and position with zero velocity.
The specific datasets we run on for the TUM-VI are the room1, room3, and room5.
For the Vicon room dataset, the groundtruth has been generated using the vicon2gt utility \cite{Geneva2020TRVICON2GT}.
The shorter TUM-VI dataset has more time periods where multiple robots are looking at the same environmental location ($26.7\%$ and $41.8\%$ of the frames detected common features without and with loop-closure), thus provides a good insight into an expected performance in a multi-user AR case where many users are observing the same environment at the \textit{same time}.
On the other hand, the Vicon room dataset has near-zero time periods where we are able to detect common features between robots by matching the most recent features.
Thus, we use the Vicon room dataset to show the accuracy gain from leveraging historical loop-closure information by matching to historical states ($28.8\%$ of the frames detected common loop-closure features).

\subsection{TUM-VI Dataset}

We use a sliding window of 11, a max of 5 SLAM features, max 30 VIO features per update, 300 active tracks, and perform online calibration of all parameters.
For the historical method, we insert keyframes into our database at 5Hz and detect and match to historical keyframes at each timestep.
We used a static weight of $\omega_i=0.99$ and distribute the remaining weight to all other robot covariances used in the CI-EKF update, and for constraint measurement updates [see Eq. \eqref{eq:constraint_res}], we used a value of $\omega_i=0.995$ and injected a synthetic measurement noise of 2cm to relax the hard constraint.

The Relative Pose Error (RPE) \cite{Zhang2018IROS} results are shown in Tab.~\ref{tab:exp_tum} solidify the performance gains due to leveraging common features from other robot agents.
The independent methods which leverage only independent VIO and SLAM feature updates have about three times the error compared to the distributed method which leverages loop-closure information.
Additionally, we can see that all variations which leverage common features are able to reduce errors due to the additional information.
It is also important to note that even though the distributed variants do not track the cross-covariances between robotic states, the use of CI allows the accuracy to be near the same level as that of the centralized algorithm, and in the case where we leverage historical information (which the centralized algorithm is unable to do), we can slightly outperform for longer trajectory length.
The \texttt{dc-full-history} method, which leverages loop-closure information, has a relatively constant error as the trajectory lengths increase as expected (showing its drift-free nature).

\subsection{Vicon Room Dataset}

We now present results on the longer hand-held, approximately 500 meter and 10 minute trajectory.
We use a sliding window of 11, a max of 20 SLAM features, max 30 VIO features per update, 200 active tracks, and perform online calibration of all parameters.
The RPE results for different segment lengths can be found in Tab. \ref{tab:exp_cyclops} and give the same conclusion as the previous TUM-VI dataset.
It is also important to note that there is very similar performance of the \texttt{indp-slam} and \texttt{ce-cmsckf-cslam} methods (and their distributed equivalents).
This is expected as there are no time periods in any of the robotic trajectories where robots are looking at the same location at the \textit{same} time.
Compared to these cases, we have huge accuracy gains due to the inclusion of common feature measurement constraints in the historical case, with halved orientation errors and a quarter of the position error at long trajectory lengths.
We also plot the groundtruth, \texttt{indp-slam}, and \texttt{dc-full-history} Robot 0 trajectories in Fig. \ref{fig:loop_path}, which reinforces that by leveraging historical information we are able to prevent inherent drift in the loop-closure-free case.

 \section{Conclusions and Future Work}

In this work we have presented a distributed  visual-inertial  cooperative CL estimator that efficiently fuses constraints between robots  and leverages temporal SLAM and loop-closure information.
We have introduced two different ways to incorporate temporal SLAM features: 
(i) directly update using the other robot's measurements, 
and (ii) if both robots are estimating the SLAM feature, a constraint between the two feature positions is leveraged.
We have adapted CI to ensure consistent fusion of loop-closure constraints to other agent's historical poses and SLAM features whose cross-correlations are unknown.
Extensive simulation and real-world evaluations have demonstrated the performance of the proposed method in realistic scenarios and showed impressive accuracy gains over the single robot case.
In the future we will focus on the practical  deployment to a low-cost low-power multi-robot application and incorporate relative robot-to-robot measurements to further increase accuracy gains.

{
\def\bibfont{\tiny}
\tiny
\bibliographystyle{IEEEtran}
\bibliography{library/rpng, library/ucr, library/related, library/library}
}

\end{document}